\documentclass[10pt,twocolumn,letterpaper]{article}

\usepackage{iccv}
\usepackage{times}
\usepackage{epsfig}
\usepackage{graphicx}
\usepackage{amsmath}
\usepackage{amssymb}
\usepackage{multirow}
\usepackage{multicol}
\usepackage{booktabs}
\usepackage{pifont}
\usepackage{makecell}
\usepackage[table]{xcolor}
\usepackage{tikz}
\usepackage[accsupp]{axessibility}

\newcommand{\cmark}{\ding{51}}%
\usepackage[pagebackref=true,breaklinks=true,letterpaper=true,colorlinks,bookmarks=false]{hyperref}

\newif\ifdraft
\draftfalse
\drafttrue

\usepackage{xcolor}
\definecolor{orange}{rgb}{1,0.5,0}
\definecolor{violet}{RGB}{150,0,170}

\definecolor{DarkMagenta}{rgb}{0.7, 0.0, 0.7}
\definecolor{DarkBlue}{rgb}{0.0, 0.0, 0.8}
\definecolor{DarkOrange}{rgb}{1.0, 0.55, 0.0}
\definecolor{DarkGreen}{rgb}{0.0, 0.5, 0.0}
\definecolor{Random}{rgb}{0.1, 0.5, 0.9}

\ifdraft
 \newcommand{\YH}[1]{{\color{Random}{\bf YH: #1}}}
 
 \newcommand{\Yang}[1]{{\color{violet}{\bf Yang: #1}}}

\else
 \newcommand{\MS}[1]{}
 
 \newcommand{\YH}[1]{}
 
 \newcommand{\Yang}[1]{}
\fi

\newcommand{\bF}{\mathbf{F}}

\newcommand{\bK}{\mathbf{K}}
\newcommand{\bP}{\mathbf{P}}
\newcommand{\bR}{\mathbf{R}}

\newcommand{\bI}{\mathbf{I}}
\newcommand{\bu}{\mathbf{u}}

\newcommand{\bp}{\mathbf{p}}
\newcommand{\bt}{\mathbf{t}}

\iccvfinalcopy 

\def\iccvPaperID{8656} 

\begin{document}

\title{Pseudo Flow Consistency for Self-Supervised 6D Object Pose Estimation}

\author{
{Yang Hai $^1$, \quad Rui Song $^1$, \quad Jiaojiao Li $^1$, \quad David Ferstl $^2$, \quad Yinlin Hu $^2$}\\
{\small $^1$ State Key Laboratory of ISN, Xidian University, \quad $^2$ MagicLeap} \\
}


\maketitle

\begin{abstract}

Most self-supervised 6D object pose estimation methods can only work with additional depth information or rely on the accurate annotation of 2D segmentation masks, limiting their application range. In this paper, we propose a 6D object pose estimation method that can be trained with pure RGB images without any auxiliary information. We first obtain a rough pose initialization from networks trained on synthetic images rendered from the target's 3D mesh. Then, we introduce a refinement strategy leveraging the geometry constraint in synthetic-to-real image pairs from multiple different views. We formulate this geometry constraint as pixel-level flow consistency between the training images with dynamically generated pseudo labels. We evaluate our method on three challenging datasets and demonstrate that it outperforms state-of-the-art self-supervised methods significantly, with neither 2D annotations nor additional depth images.

\end{abstract}

\section{Introduction}

The goal of 6D object pose estimation is to accurately estimate the 3D rotation and 3D translation of a rigid object with respect to the camera, which gives essential information about the world beyond classical 2D understanding and is a fundamental component in many applications, such as robotic manipulation~\cite{collet2011moped}, autonomous driving~\cite{manhardt2019roi}, and augmented reality~\cite{marchand2015pose}.

Recent progress in this field has significantly improved the robustness and accuracy of the model~\cite{ZebraPose_2022_cvpr, xu2022rnnpose, Repose_2021_iccv, Gdr-net_2021_cvpr, So-Pose_2022_iccv, DeepIM_2018_eccv, WDR_2021_cvpr, Segdriven_2019_cvpr}. Most of these approaches, however, rely on a large number of real images with accurate 6D pose annotations. But, compared to classical 2D annotation, these 6D annotations are either very hard to obtain~\cite{marion2018label, liu2021stereobj} or are prone to contain large labeling errors~\cite{PFA_2022_eccv, surfemb_2022_cvpr, PoseCNN_2018_rss}.

\begin{figure}
    \centering
    \begin{tabular}{c}
    \includegraphics[width=0.95\linewidth]{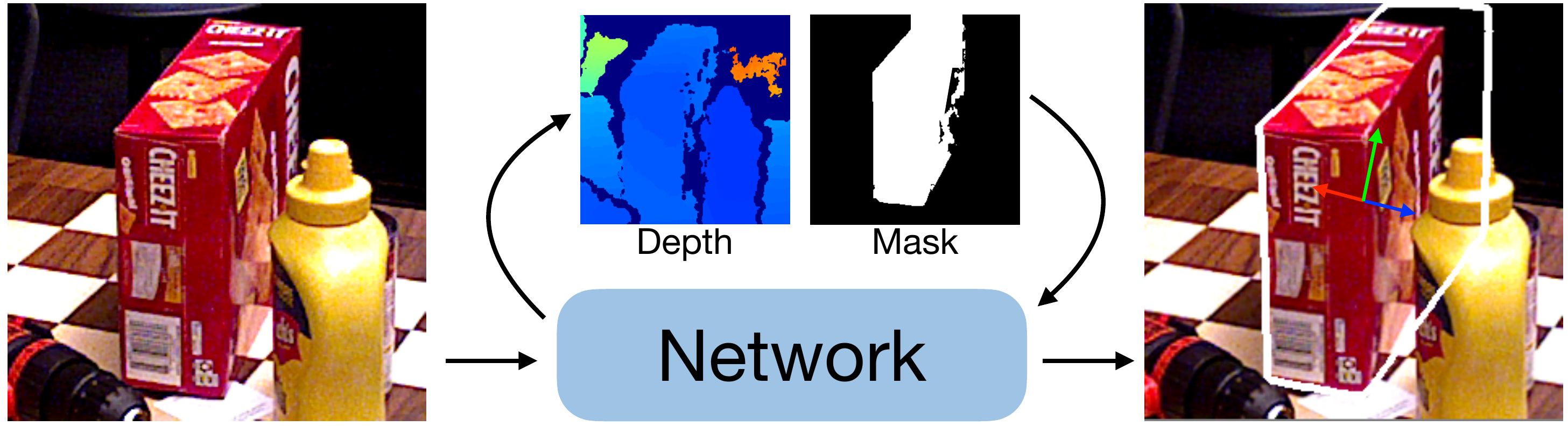} \\
    \small (a) Existing self-supervised 6D object pose methods\\
    \includegraphics[width=0.95\linewidth]{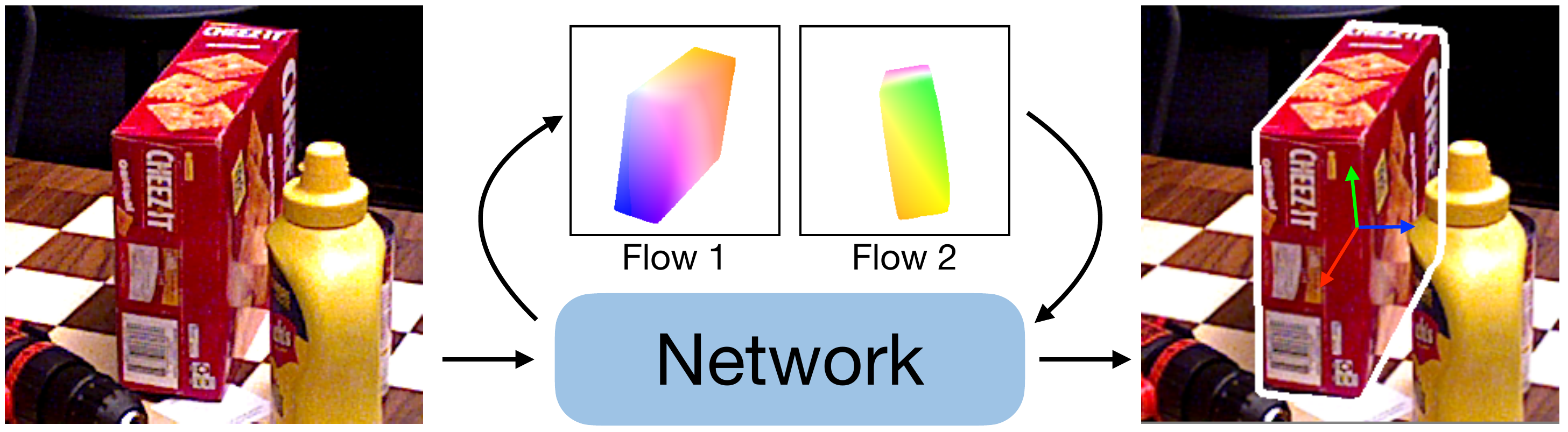} \\
    \small (b) Our solution based on pseudo flow consistency  \\
\end{tabular}
    \caption{{\bf Comparison of self-supervised object pose methods.}
    {\bf (a)} Most existing self-supervised object pose methods rely on either the depth image~\cite{Self6d_2020_eccv, Self6Dpp_2021_tpami, chen2022sim, lin2022learning} or additional mask annotations~\cite{DSCPose_2021_cvpr, sock2020introducing}, limiting their application range.
    {\bf (b)} By contrast, our method can be trained only with the guidance of flow consistency based on the intrinsic geometry constraint of multiple different views, and produces more accurate results than existing solutions, without relying on any auxiliary information.}
    \label{fig:teaser}
\end{figure}

Some recent methods propose to use techniques based on image synthesis~\cite{hodan2018bop} or self-supervised learning~\cite{Self6d_2020_eccv, Self6Dpp_2021_tpami} to handle this problem. The main problem with synthetic images is the large domain gap to the real images, making the model's generalization ability suffers in practice~\cite{PVNet_2019_cvpr, xu2022rnnpose}. On the other hand, most self-supervised methods rely on additional information. Some can only work with additional depth images~\cite{Self6d_2020_eccv, Self6Dpp_2021_tpami, lin2022learning, chen2022sim} or others need pixel-level annotation of a segmentation mask~\cite{sock2020introducing, DSCPose_2021_cvpr}, which prevents the general applicability, as shown in Fig.~\ref{fig:teaser}.

\begin{figure*}
\centering
    \setlength\tabcolsep{12pt}
    \begin{tabular}{cc}
        \includegraphics[width=0.40\linewidth] {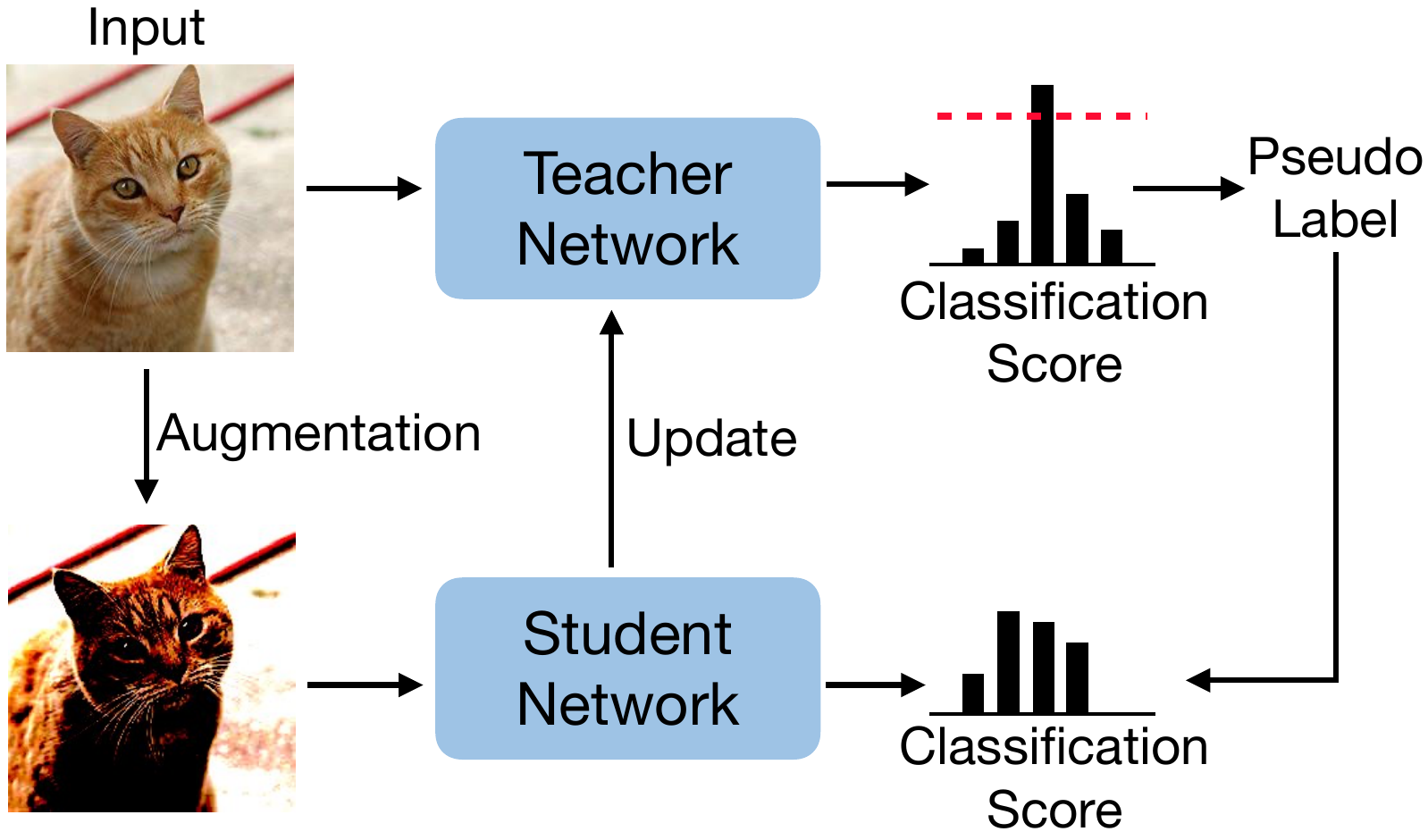} &
        \includegraphics[width=0.48\linewidth]{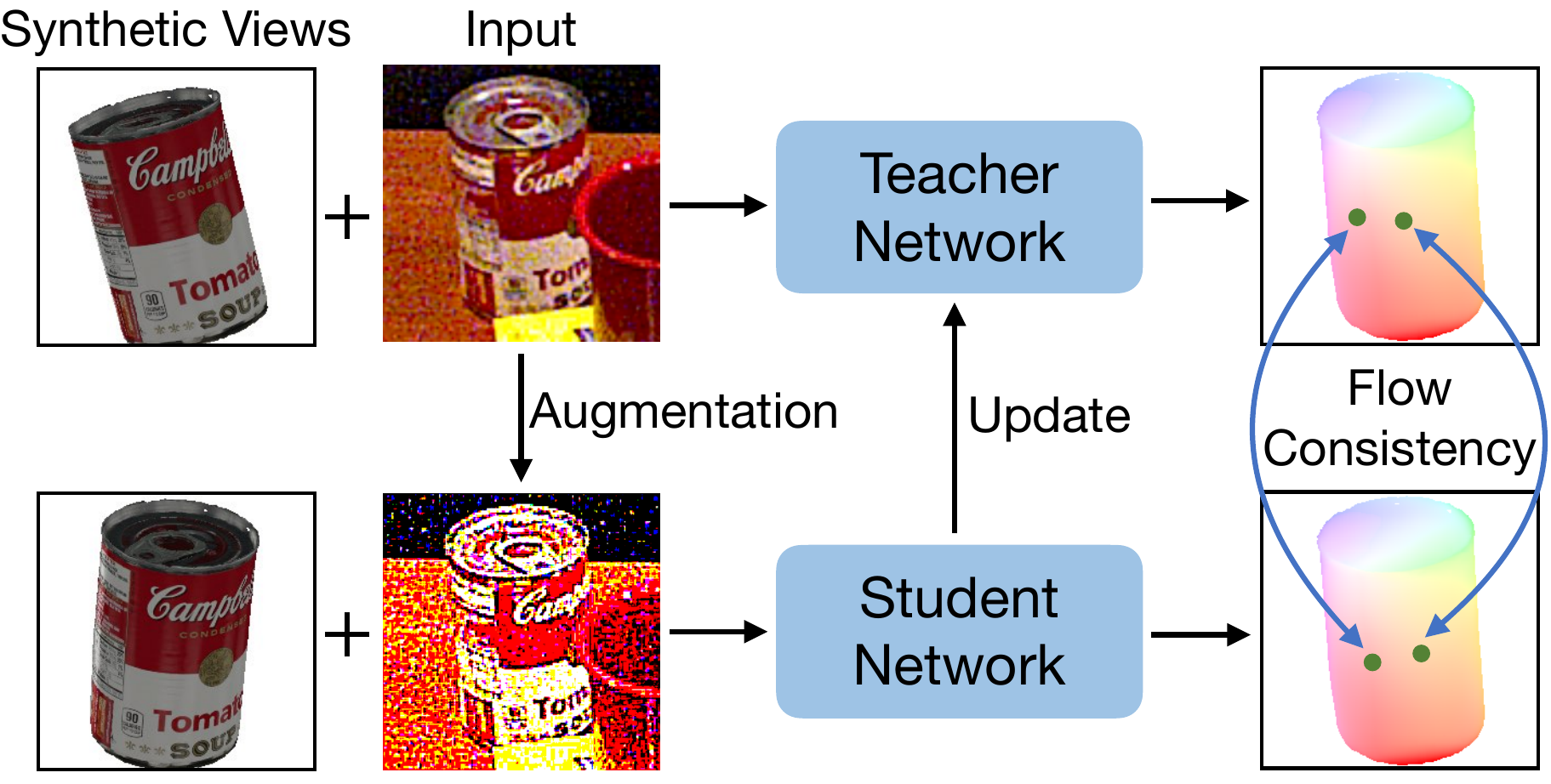} \\
        {\small (a) The standard strategy for self-supervised classification} &
        {\small (b) The proposed strategy for self-supervised object pose estimation} \\
    \end{tabular}
    \vspace{0.5em}
    \caption{{\bf Self-supervised strategies in different fields.}
        {\bf (a)} Teacher-student learning scheme is a classical framework for self-supervised classification~\cite{tarvainen2017mean}. The key is how to determine the quality of pseudo labels from the noisy prediction of the teacher network. For image classification, one can obtain the prediction quality by the output distribution after the softmax operation easily, which is usually implemented by checking if the probability of any class is above a threshold~\cite{sohn2020fixmatch, zhang2021flexmatch}.  {\bf (b)} However, there is no such easy way to determine the quality of an object pose prediction without the ground truth. We propose to formulate pseudo object pose labels as pixel-level optical flow supervision signals, and then use the flow consistency between multiple views based on their underlying geometry constraint.}
    \label{fig:framework_compare}
\end{figure*}


In this work, we propose a self-supervised framework for 6D object pose estimation, which relies on neither depth nor additional 2D annotations. We first generate a synthetic dataset based on rendered images from the target's 3D mesh and train networks only on this dataset to get a rough pose initialization. To close the domain gap between the synthetic and real data, we use a refinement strategy where we compare the rendered reference image according to the initial pose and the real input based on pseudo labels~\cite{sohn2020fixmatch, zhang2021flexmatch}. Pseudo labeling is widely used in many computer vision tasks~\cite{tarvainen2017mean, xu2021softteacher, he2020moco, grill2020byol}. However, the two fundamental problems of pseudo labeling are still open questions in 6D object pose estimation, including the generative strategy of creating pseudo labels and the selection strategy of extracting high-quality labels from the noisy candidates, as shown in Fig.~\ref{fig:framework_compare}.

We propose to formulate the pseudo 6D pose labels as pixel-level flow supervision signals in a render-and-compare framework~\cite{PFA_2022_eccv, cosypose_2020_eccv, DeepIM_2018_eccv, xu2022rnnpose, Repose_2021_iccv, coupled_2022_cvpr, yang2023shapeflow}. Unlike the common render-and-compare frameworks that need accurate pose annotations, we propose a geometry-guided learning framework without any annotations. We render multiple images near the initial pose, and compare them with the real input with the guidance of flow consistency, based on the geometry constraints between these image pairs from different views. We choose high-quality flow labels based on the proposed consistency on the fly, and supervise the network training with these dynamically generated labels in every training step.

We evaluate our method on three challenging datasets LINEMOD~\cite{Linemod_2012_accv},  Occluded-LINEMOD~\cite{OccLinemod_2015_iccv}, and YCB-V~\cite{PoseCNN_2018_rss}, and show that it outperforms state-of-the-art self-supervised methods significantly, including those methods relied on depth image~\cite{Self6d_2020_eccv, Self6Dpp_2021_tpami} or auxiliary annotation information~\cite{DSCPose_2021_cvpr, sock2020introducing}.

Our contributions can be summarized as the following. First, we investigate the problem of the standard teacher-student methods in selecting high-quality pseudo labels for self-supervised object pose estimation. Second, we propose a strategy based on flow consistency that embeds the geometry constraint from multiple views. Finally, we demonstrate its effectiveness by significantly outperforming state-of-the-art self-supervised object pose methods, without relying on any auxiliary information.

\begin{figure*}
    \centering
    \includegraphics[width=0.99\linewidth]{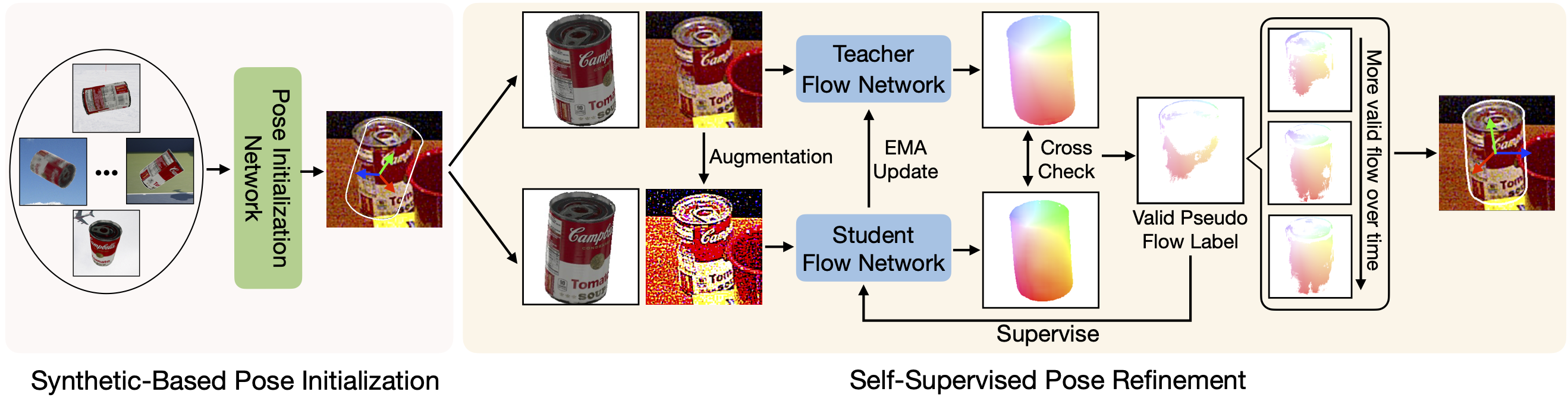}
    \caption{{\bf Method overview.}
        We first obtain the initial pose based on a pose estimation network trained only on synthetic images, and then train our refinement framework on real images without any annotations. Our proposed framework is based on a teacher-student learning scheme. Given a rough pose initialization, we render multiple synthetic images around this initial pose, and create multiple image pairs between the synthetic and real images.
        We dynamically produce pixel-level flow supervision signals for the student network during the training, by leveraging the geometry-guided flow consistency between those image pairs from different views. After getting 3D-to-2D correspondences based on the predicted flow, we use a PnP solver to get the final pose~\cite{PFA_2022_eccv}.
    }
    \label{fig:arch}
\end{figure*}

\section{Related Work}

Object pose estimation has shown significant progress recently, based on different techniques, such as direct pose regression~\cite{Gdr-net_2021_cvpr, So-Pose_2022_iccv, cosypose_2020_eccv}, 2D reprojection regression~\cite{rad2017bb8, PVNet_2019_cvpr, Segdriven_2019_cvpr, WDR_2021_cvpr, SuoSLAM_2022_cvpr}, 3D keypoint prediction~\cite{CDPN_2019_iccv, pix2pise_2019_iccv, ZebraPose_2022_cvpr, surfemb_2022_cvpr}, and differentiable PnP solver~\cite{EPnP_2009_ijcv, Single-stage_2020_cvpr, chen2020end, chen2022epro}.
However, most of these methods rely on a large number of real images with accurate 6D pose annotation, which is usually hard to obtain in practice, especially in cluttered scenes with multiple object instances and occlusions~\cite{PoseCNN_2018_rss, PFA_2022_eccv}. 

Some recent methods tackle this problem by training on synthetic images rendered from the target's 3D mesh~\cite{surfemb_2022_cvpr, PVNet_2019_cvpr, sc6d_3dv_2022}, but this strategy suffers from the domain gap between the synthetic and real image sets~\cite{ denninger2020blenderproc}.
In contrast, some pose refinement methods have shown significant improvement in the generalization ability across different domains~\cite{DeepIM_2018_eccv, coupled_2022_cvpr, cosypose_2020_eccv, Repose_2021_iccv, xu2022rnnpose, yang2023shapeflow} and especially~\cite{PFA_2022_eccv}, which produces comparable results as the state-of-the-art methods with only about one-tenth of the real images involved in training. Although having this promising progress, these pose refinement methods still need many annotated real images for training, and can not easily benefit from more real data without further annotations.

To solve this problem, some recent self-supervised methods~\cite{Self6d_2020_eccv, DSCPose_2021_cvpr, sock2020introducing, Self6Dpp_2021_tpami} try to remove the cumbersome procedure of pose annotation completely. Most of them are based on a strategy that compares the synthetic image rendered from an initial pose with the real image, and backpropagate the gradient through a differentiable renderer~\cite{neuralrender_2018_cvpr, softrasterizer_2019_iccv} to update the network's weights during training, expecting to align the rendered image with the real input without explicit annotations. This type of strategy, however, relies heavily on the performance of comparing the final rendered image and the real input, which suffers from the domain gap, making them often rely either on depth ~\cite{Self6d_2020_eccv} or on additional pixel-level annotations of segmentation masks~\cite{DSCPose_2021_cvpr}. In contrast, we propose to compare multiple synthetic-to-real image pairs at the same time, and force networks to comply with the geometry constraint between those image pairs from different views, which suffers little from the domain problem. On the other hand, we formulate the geometry constraint as pixel-level consistency, which provides us dynamic valid label mask during training, without any 3D depth information or additional 2D mask annotations.

Pseudo labeling is also one of the basic techniques used in recent self-supervised object pose methods~\cite{lin2022learning, chen2022sim}. However, these approaches still rely on additional depth images to select valid pseudo labels. In addition, they only update the pseudo labels after finishing the previous training, which usually means the model needs to be trained multiple times to utilize the slowly updated pseudo labels. By contrast, our pseudo flow labels are generated dynamically in every training step, and our model only needs to be trained once. 

Our method is related to the recent teacher-student formulation of pseudo labeling~\cite{sohn2020fixmatch, zhang2021flexmatch, tarvainen2017mean, xu2021softteacher, kwon2022semi, zhou2022dense, li2022pseco, Zhang_2018_CVPR, wu2023spatiotemporal, qiu2023coupling, katircioglu2021self}, which works under the assumption that the generated high-quality pseudo label of the teacher can be used to supervise the student network when having the same input as the teacher but only different data augmentations. Although this simple general framework has been widely used in image classification~\cite{sohn2020fixmatch, zhang2021flexmatch}, object detection~\cite{xu2021softteacher, li2022pseco, zhou2022dense, tang2021humble}, and semantic segmentation~\cite{kwon2022semi}, it only can work with high-quality pseudo labels. However, there is still no easy way to generate high-quality pose labels in the context of 6D object pose estimation. To solve this problem, we propose to formulate pseudo 6D pose labels as pixel-level flow supervision signals and select high-quality pseudo flow labels based on flow consistency across multiple different views during training. Our experiments demonstrate the effectiveness of this method.

\section{Approach}

Given a dataset of calibrated RGB images and the 3D mesh of the target, our goal is to train a self-supervised model on this dataset to estimate the 6D object pose of the target, without relying on depth images or any auxiliary information, such as 6D pose and 2D mask annotations. We first create a synthetic dataset by rendering the 3D mesh of the target in different poses and train an existing pose estimation network~\cite{WDR_2021_cvpr, cosypose_2020_eccv} on it to obtain a rough pose initialization~\cite{PFA_2022_eccv,DeepIM_2018_eccv,Repose_2021_iccv}. The core component of our method is a self-supervised pose refinement framework, which we will discuss in detail in this section. We first show an overview of our self-supervised framework, and then present how we formulate the flow consistency based on the geometry constraint between different views. Finally, we show how we extend it to multiple image pairs to further increase the robustness. Fig.~\ref{fig:arch} shows the overview of our method.

\begin{figure*}
    \centering
    \setlength\tabcolsep{15pt}
    \begin{tabular}{cc}
        \includegraphics[width=0.45\linewidth]{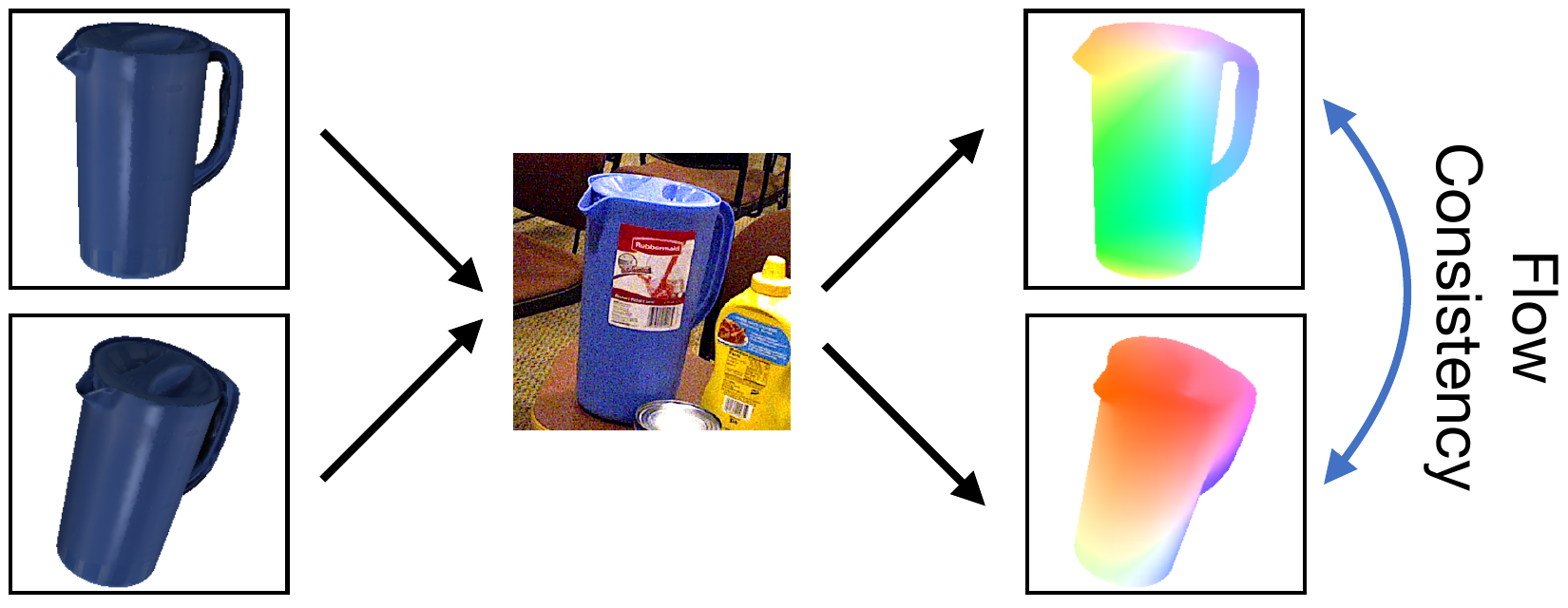} &
        \includegraphics[width=0.45\linewidth]{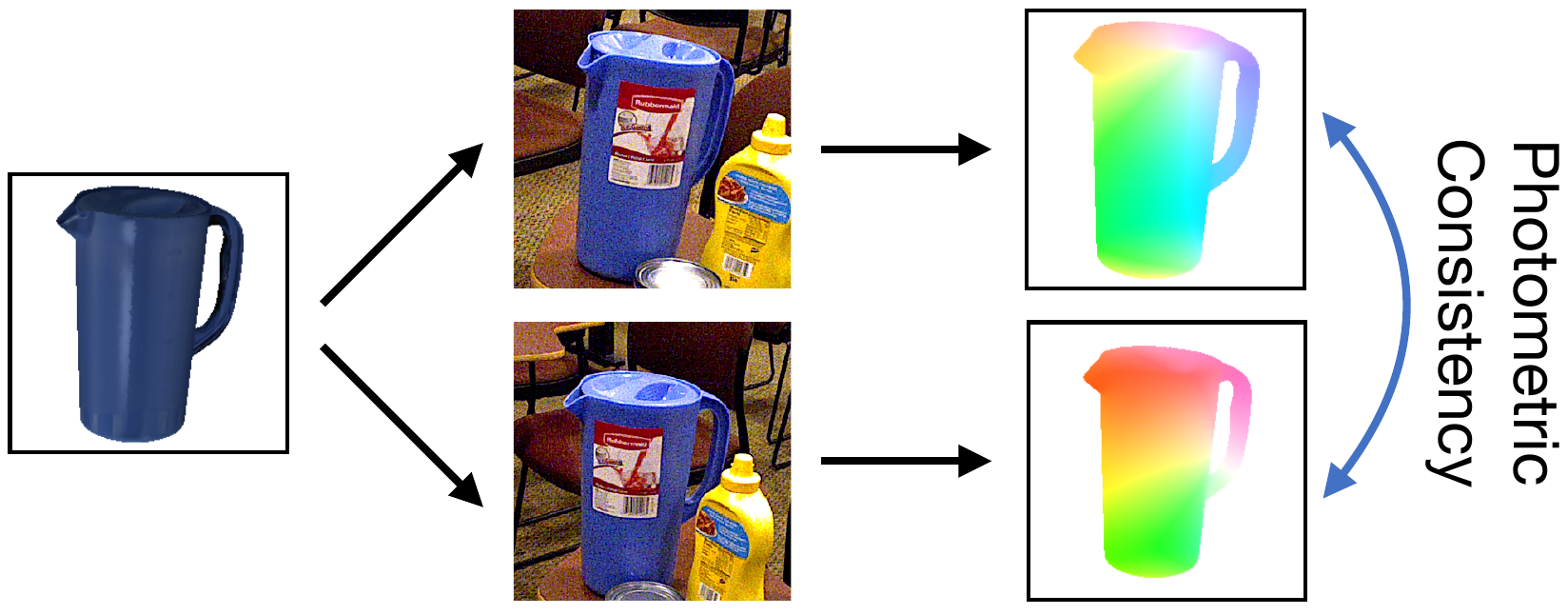} \\
        {\small (a) Flow consistency across multiple views} &
        {\small (b) Flow-guided photometric consistency} \\
    \end{tabular}
    \vspace{0.2em}
    \caption{{\bf Illustration of geometry-guided consistency.}
    We predict the flow between synthetic and real images. {\bf (a)} The 2D flow of the same 3D point from different synthetic views to the real input should be consistent.
    {\bf (b)} On the other hand, the target 2D locations of the same 3D point on different real inputs should have similar textures.
    }
    \label{fig:flow_consistency}
\end{figure*}

\subsection{Framework Overview}
\label{sec:overall}

We use a teacher-student architecture~\cite{tarvainen2017mean} for our self-supervised framework. It contains two networks with identical network structures, but not shared weights, which are called the teacher and student, respectively. During training, when an image input of the teacher network can produce a prediction that can fulfill some criteria, we convert this prediction to a one-hot pseudo-label, and use it to supervise the student network with the same image input but only different data augmentations. After updating the weights of the student network by gradient backpropagation supervised by this pseudo label, we then update the weights of the teacher network by a simple exponential moving averaging (EMA) strategy from the student network: 
\begin{equation}
{\bf W}_t = \alpha {\bf W}_t + (1-\alpha){\bf W}_s, 
\end{equation}
where $ {\bf W}_t $ and ${\bf W}_s$ are the network weight parameters of the teacher and student network, respectively, and $\alpha$ is the exponential factor, which is typically 0.999. The weight updating and pseudo label generation is conducted after each iteration during training, making it much more efficient than other pseudo-label-based object pose methods~\cite{chen2022sim, lin2022learning}, which can only produce pseudo labels after the whole training pipeline and need to train the model multiple times. 

Our main problem is how to select high-quality pseudo label candidates from the noisy predictions of the teacher network. For the image classification task as in~\cite{tarvainen2017mean}, one can obtain the label quality by the output distribution after the softmax operation easily, which is usually implemented by checking if the probability of any class is above a threshold~\cite{sohn2020fixmatch, zhang2021flexmatch}. However, there is no such easy way to determine the quality of an object pose prediction without the ground truth. We discuss our solution in the following sections.

\subsection{Flow Consistency across Multiple Views}
\label{sec:geo_con}

To solve the problem of difficulties in determining the quality of object pose predictions, we first formulate object pose estimation as a problem of estimating dense 2D-to-2D correspondence, or optical flow estimation, as in PFA~\cite{PFA_2022_eccv}, which, however, is a fully-supervised object pose method. To tackle the problem of no pose annotation for the computation of ground truth flow, we render multiple images around the initial pose, and predict the flow between each of them and the real input. In principle, since both the rendered images and real input image are 2D reprojections of the same 3D object, the flow prediction that aligns with the underlying geometry should have a higher probability of being of high quality. Fig.~\ref{fig:flow_consistency}(a) illustrates such consistency assumption. 

More formally, given an unannotated real image $\bI^t$ and the obtained initial pose $\bP_0$ from networks trained only on synthetic data, we randomly generate another $n-1$ poses $\{\bP_1, \cdots, \bP_{n-1}\}$, around the initial pose $\bP_0$, and then create $n$ synthetic images by rendering the target under the corresponding poses, generating $n$ image pairs:
\begin{equation}
    \{(\bI_i^r, \bI^t)\}, \ \ \ \  0 \leq i \leq n-1,
    \label{eq:image_pairs1}
\end{equation}
where $\bI_i^r$ is the rendered image of the target under pose $\bP_i$.

For an object having $N$ 3D keypoints, the 2D reprojection of a 3D keypoint $\bp_j, 1 \leq j \leq N,$ under pose $\bP_i$ can be obtained by
\begin{equation}
    \begin{aligned}
        \lambda_{ij}^r
        \begin{bmatrix}
        \bu_{ij}^r \\
        1
        \end{bmatrix}
         = \bK(\bR_i \bp_j + \bt_i),
    \end{aligned}
    \label{eq:perspective}
\end{equation}
where $\lambda_{ij}^r$ is a scale factor, $\bu_{ij}^r$ is the 2D image location, $\bK$ is the intrinsic camera matrix, and $\bR_i$ and $\bt_i$ are the rotation and translation of pose $\bP_i$, respectively. We then establish 3D-to-2D correspondence ${\bf p}_j \leftrightarrow {\bf u}_{ij}^r$ under pose $\bP_i$.
For the real image $\bI^t$ , although its true pose $\bP^t$ is unknown to us, the relation between the 3D keypoint $\bp_j$ and its 2D image location $\bu_j^t$ should follow the perspective principle of Eq.~\ref{eq:perspective}, implicitly generating the correspondence ${\bf p}_j \leftrightarrow \bu_j^t$.

We train a network to predict dense 2D-to-2D correspondence $\bF^{r \rightarrow t}_i$ between the two images in each image pair of Eq.~\ref{eq:image_pairs1}, such that
\begin{equation}
\bu_{ij}^r + {\bf f}_{i}^{r \rightarrow t} = \bu_{ij}^t \; ,
\label{eq:flow_corr}
\end{equation}
where ${\bf f}_{i}^{r \rightarrow t} $ is the corresponding 2D flow vector.
Although $\bu_j^t$ is unknown during training, we have the geometry constraint that the 2D image locations $\{\bu_{ij}^r + {\bf f}_{i}^{r \rightarrow t}\}$ of the same 3D keypoint $\bp_j$ from different synthetic views $0 \leq i \leq n-1$ should be the same.

We use the standard variance of the predicted $\bu_{ij}^t$ from different views to determine if the current pixel's flow prediction is a valid pseudo label
\begin{equation}
    \sigma_{j} = std(\{\bu_{ij}^r + {\bf f}_{i}^{r \rightarrow t}\}) \quad 0 \leq i \leq n-1\;, 
\end{equation}
and select valid flow pseudo labels by a simple threshold $\tau$. Fig.~\ref{fig:training_sigma} shows some visualizations of the variance $\sigma$.

After obtaining the valid flow labels from the teacher network, we use them to supervise the student network by a loss function
\begin{equation}
    \mathcal{L}_{flow} = \sum_{i=0}^{n-1} V_i \Vert (g(\bI_i^r, \bI^t; {\bf W}_t) - g(\bI_i^r, \tilde{\bI}^t; {\bf W}_s)) \Vert \; ,
\end{equation}
where $g$ is the flow network with parameters $ {\bf W}_t $ and ${\bf W}_s $ for the teacher and student network, respectively, and $\tilde{\bI}^t$ is the same real image as $\bI^t$ but only with different data augmentations, and $V_i$ is the mask containing valid pixels where $\sigma_{j}<\tau$. 
Note that, $V_i$ is generated dynamically from the consistency check between multiple image pairs in Eq.~\ref{eq:image_pairs1}, and does not rely on any 2D mask annotations.

\subsection{Flow-Guided Photometric Consistency}
\label{sec:photo_con}

The previous section only investigates the consistency between the synthetic views and the real input. We further explore the consistency between multiple real inputs.
Our motivation is that the 2D image reprojections of the same 3D object keypoint on different real images should have similar textures. We formulate this texture assumption as a photometric consistency, as illustrated in Fig.~\ref{fig:flow_consistency}(b).

Given the real image ${\bI}^t$ with the initial pose ${\bP_0}$, as in the previous section, we randomly retrieve another $m$ real images whose initial pose is around $\bP_0$, generating $m$ image pairs
\begin{equation}
    \{(\bI_0^r, \bI^t_k)\}, \ \ \ \  1 \leq k \leq m,
    \label{eq:image_pairs2}
\end{equation}
and after feeding the two images in each image pair to the teacher network, we have
\begin{equation}
{\bar\bu}_k^t = \bu_{0}^r + g(\bI_0^r, \bI_k^t; {\bf W}_t) \; ,
\label{eq:flow_corr_2}
\end{equation}
where ${\bar\bu}_k^t$ is the predicted 2D image location on image $\bI^t_k$.
We assume these predicted 2D image locations of the same 3D keypoint have similar texture properties, and we use a photometric loss to model this
\begin{equation}
    \mathcal{L}_{photo} = \sum_{k=1}^{m} V_0 \rho(w(\bI_k^t, {\bar\bu}_k^t), w(\bI^t, {\bar\bu}^t)) \; ,
\end{equation}
where $w$ is an operation function that warps the image according to the new pixel locations, $\rho$ is a generalized Charbonnier function to measure the photometric difference based on the Census transformation~\cite{meister2018unflow}, and ${\bar\bu}^t$ is inferred from the student's prediction, where
\begin{equation}
    {\bar\bu}^t = \bu_{0}^r + g(\bI_0^r, \tilde{\bI}^t; {\bf W}_s)
\end{equation}

We combine the flow consistency and photometric consistency into our final loss
\begin{equation}
    \mathcal{L} = \mathcal{L}_{flow} + 0.5\mathcal{L}_{photo}.
\end{equation}
Note that, we only apply the loss to the student network since the gradient backprogataion only occurs for the student network, and the teacher network only gets its weight updated by EMA updating as discussed in Section~\ref{sec:overall}.

\section{Experiments}
In this section, we first present the experiment setting of our method and then compare our method with state-of-the-art self-supervised methods. We finally conduct detailed ablation studies of our method in various settings. Our source code is publicly available at \url{https://github.com/YangHai-1218/PseudoFlow}.

\subsection{Experiment Setup}

\noindent \textbf{Datasets.}
We evaluate our method on three widely-used datasets for 6D object pose estimation: LINEMOD (``LM'')~\cite{Linemod_2012_accv}, Occluded-LINEMOD (``LM-O'')~\cite{OccLinemod_2015_iccv}, and YCB-V~\cite{PoseCNN_2018_rss}. LINEMOD dataset contains 13 objects, with a single sequence per object without occlusions. We follow ~\cite{Self6d_2020_eccv, PFA_2022_eccv} to use 15\% of the real images for training, resulting in a total of 2.4k images. Occluded-LINEMOD is an extension of LINEMOD, which annotates all the objects in one sequence in LINEMOD as the test set and shares the training set with LINEMOD. The recent YCB-V dataset consists of 130k real training images for 21 texture-less objects captured in cluttered scenes. 
Although all these three datasets contain manually labeled annotations, we train our models on them without accessing the ground truth, and report the final accuracy on their test set.
We use the synthetic dataset used in the BOP challenge~\cite{denninger2020blenderproc, hodan2018bop, sundermeyer2023bop} to train WDR-Pose~\cite{WDR_2021_cvpr} for the pose initialization.

\noindent \textbf{Evaluation Metrics.}
We mainly use ADD-0.1d as our metric, which computes the average distance between the mesh vertices transformed by the predicted pose and the ground truth pose, and then only treat the prediction with an average 3D error below 10\% of the mesh diameter as a correct pose estimate. We use its symmetric version for symmetric objects.
Additionally, in some settings, we also use BOP metrics~\cite{hodan2018bop} for evaluation, including the Visible Surface Discrepancy (VSD), the Maximum Symmetry-aware Surface Distance (MSSD), the Maximum Symmetry-aware Projection Distance (MSPD), and their average AR. We refer the readers to~\cite{hodan2018bop} for their detailed definition.

\noindent \textbf{Training details.}
We use RAFT~\cite{RAFT_2020_eccv} as our flow network for both the teacher and student network. We initialize the weights of both teacher and student network with the weights pretrained on synthetic data and train the model using AdamW optimizer~\cite{AdamW_2019_iclr} with a batch size of 16.
We use One-cycle strategy~\cite{smith2019super} to anneal the learning rate from a starting point 4e-4.
We crop the target object from the original image according to the initial pose, and then resize the image patch to 256$\times$256. We do not use any data augmentation in the teacher network, and only use random color augmentation used in PFA~\cite{PFA_2022_eccv} for the student network.
We typically set $\tau=1$, $m=3$, and $n=4$ in our experiments.
Unlike~\cite{Self6d_2020_eccv, Self6Dpp_2021_tpami, lin2022learning, huang2022neural} that train a separate model for each object, which is cumbersome to train, we train a single model for all objects in the same dataset.

\subsection{Comparison against State of the Art}
We first compare our method against the state-of-the-art self-supervised pose estimation methods on LINEMOD and Occluded-LINEMOD.
Since most of them report numbers only in ADD-0.1d, we follow the same for a fair comparison. Table.~\ref{tab:compare_lm} and~\ref{tab:compare_lmo} summarize the result.
Our method outperforms the state-of-the-art methods significantly.
Especially, our method, which requires only RGB images, even outperforms Self6D$\ddagger$~\cite{Self6Dpp_2021_tpami} by 3.7\% on LINEMOD, which is a method that relies on additional depth images. We show some qualitative results in Fig.~\ref{fig:quantitive_reuslts}.

\begin{table}
    \centering
    \setlength\tabcolsep{1.5pt}

\begin{tabular}{l|ccccccc}
   \toprule
   Method      & \makecell{DSC\\~\cite{DSCPose_2021_cvpr}}   & \makecell{Sock et al.\\~\cite{sock2020introducing}}    
               & \makecell{Lin et al.\\~\cite{lin2022learning}}  &\makecell{Self6D\\~\cite{Self6Dpp_2021_tpami}}      & \makecell{Self6D$\ddagger$\\~\cite{Self6Dpp_2021_tpami}}  
               & {\bf Ours}  \\
   \midrule
   Ape         & 31.2      & 37.6              & 67.5          & 76.0          & 75.4      & {\bf 81.9}             \\
   Bench.      & 83.0      & 78.6              & {\bf 99.9}    & 91.6          & 94.9      & 95.0           \\
   Cam         & 49.6      & 65.6              & 87.4          & {\bf 97.1}    & 97.0      & 94.2       \\
   Can         & 56.5      & 65.6              & 99.2          & {\bf 99.8}    & 99.5      & 96.8      \\
   Cat         & 57.9      & 52.5              & 94.3          & 85.6          & 86.6      & {\bf 95.4}      \\
   Driller     & 73.7      & 48.8              & 97.6          & 98.8          & {\bf 98.9}& 94.8      \\
   Duck        & 31.3      & 35.1              & 67.2          & 56.5          & 68.3      & {\bf 83.5}      \\
   Eggbox*     & 96.0      & 89.2              & 98.9          & 91.0          & {\bf 99.0}& 93.9      \\
   Glue*       & 63.4      & 64.5              & 96.2          & 92.2          & 96.1      & {\bf 96.5}      \\
   Holep.      & 38.8      & 41.5              & 49.9          & 35.4          & 41.9      & {\bf 84.5}      \\
   Iron        & 61.9      & 80.9              & 99.5          & {\bf 99.5}    & 99.4      & 94.9      \\
   Lamp        & 64.7      & 70.7              & {\bf 99.8}    & 97.4          & 98.9      & 94.8      \\
   Phone       & 54.4      & 60.5              & 91.5          & 91.8          & {\bf 94.3}      & 94.1      \\
   \midrule
   Avg.        & 58.6      & 60.6              & 88.4          & 85.6          & 88.5      &  {\bf 92.2}     \\
   \bottomrule
\end{tabular}
    \vspace{-0.2em}
    \caption{{\bf Comparison with self-supervised methods on LINEMOD.}
    ``*'' denotes symmetric objects. We use the latest version of Self6D++~\cite{Self6Dpp_2021_tpami}, and we only denote it as ``Self6D'' for simplicity. ``Self6D$\ddagger$'' is the version with supervision from additional depth images. Our method outperforms ``Self6D$\ddagger$'' with only RGB images.
    }
    \label{tab:compare_lm}
\end{table}

\vspace{-5pt}
\begin{table}
    \centering
    \setlength\tabcolsep{1.5pt}

\begin{tabular}{l|cccccc}
   \toprule
   Method      & \makecell{DSC\\~\cite{DSCPose_2021_cvpr}}   & \makecell{Sock et al.\\~\cite{sock2020introducing}}       
               & \makecell{Lin et al.\\~\cite{lin2022learning}} &\makecell{Self6D\\~\cite{Self6Dpp_2021_tpami}}      & \makecell{Self6D$\ddagger$\\~\cite{Self6Dpp_2021_tpami}}
               & {\bf Ours}  \\
   \midrule
   Ape         & 9.1       & 12.0      & 40.3      & 57.7      & 59.4          & {\bf 60.1}      \\
   Can         & 21.1      & 27.5      & 75.2      & 95.0      & {\bf 96.5}    & 94.2      \\
   Cat         & 26.0      & 12.0      & 35.0      & 52.6      & {\bf 60.8}    & 56.5      \\
   Driller     & 33.5      & 20.5      & 68.5      & 90.5      & {\bf 92.0}    & 89.7      \\
   Duck        & 12.2      & 23.0      & 25.7      & 26.7      & 30.6          & {\bf 30.9}      \\
   Eggbox*     & 39.4      & 25.1      & 44.7      & 45.0      & 51.1          & {\bf 58.1}    \\
   Glue*       & 37.0      & 27.0      & 60.7      & 87.1      & 88.6          & {\bf 88.9}       \\
   Holep.      & 20.4      & 35.0      & 28.0      & 23.5      & 38.5          & {\bf 44.2}      \\
   \midrule    
   Avg.        & 24.8      & 22.8      & 47.3      & 59.8      & 64.7          & {\bf 65.4}      \\
   \bottomrule
\end{tabular}
    \vspace{-0.2em}
    \caption{{\bf Comparison on Occluded-LINEMOD.}}
    \label{tab:compare_lmo}
\end{table}

\subsection{Ablation Study}

\noindent \textbf{Evaluation of different components.}
Table.~\ref{tab:compare_ssl} summarizes the results of our method with different components. The first row is the results of the standard teacher-student structure used in ~\cite{sohn2020fixmatch}. Although it has the standard EMA updating strategy, its performance is limited, mainly caused by the lacking of high-quality pseudo pose labels. After adding our flow loss and photometric loss, the performance increases significantly, which demonstrates the effectiveness of the proposed components.

\begin{figure*}
    \centering
    \setlength\tabcolsep{4pt}
    \begin{tabular}{ccc}
        \includegraphics[width=0.32\linewidth]{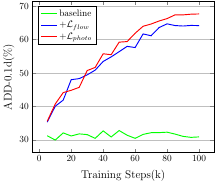} & 
         \includegraphics[width=0.31\linewidth]{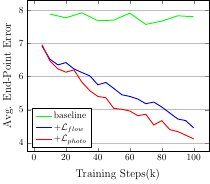} &
         \includegraphics[width=0.32\linewidth]{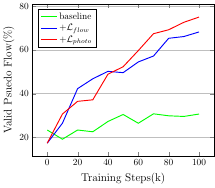} \\ 
    \end{tabular}
    \caption{{\bf Training analysis on YCB-V.}
    We report the results in three different settings during training. The baseline is the original teacher-student structure~\cite{tarvainen2017mean}, and the other two are the proposed components of our method. The baseline model struggles to learn from unannotated data. By contrast, our flow loss tackles this problem effectively, and our photometric loss increases the performance further.
    }
    \label{fig:training}
\end{figure*}

\begin{figure*}
    \centering
   \setlength\tabcolsep{1pt}
    \scalebox{0.97}{
\begin{tabular}{cccccccc}

  \includegraphics[width=0.14\linewidth]{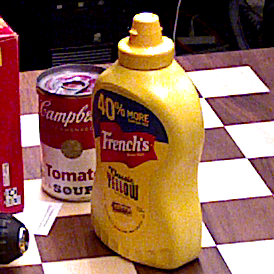} & 
  \setlength{\fboxsep}{-0.5pt}\setlength{\fboxrule}{0.5pt}\fbox{\includegraphics[width=0.14\linewidth]{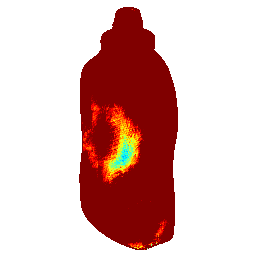}} & 
  \setlength{\fboxsep}{-0.5pt}\setlength{\fboxrule}{0.5pt}\fbox{\includegraphics[width=0.14\linewidth]{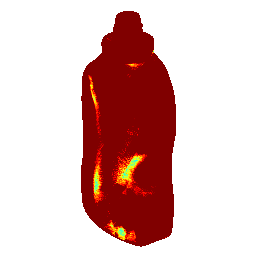}} &
  \setlength{\fboxsep}{-0.5pt}\setlength{\fboxrule}{0.5pt}\fbox{\includegraphics[width=0.14\linewidth]{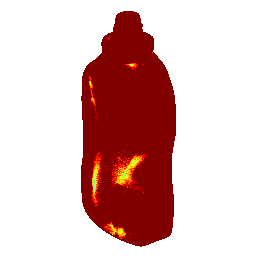}} &
  \setlength{\fboxsep}{-0.5pt}\setlength{\fboxrule}{0.5pt}\fbox{\includegraphics[width=0.14\linewidth]{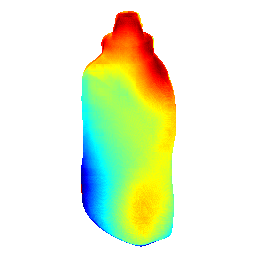}} &
  \setlength{\fboxsep}{-0.5pt}\setlength{\fboxrule}{0.5pt}\fbox{\includegraphics[width=0.14\linewidth]{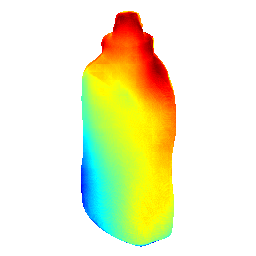}} &
  \setlength{\fboxsep}{-0.5pt}\setlength{\fboxrule}{0.5pt}\fbox{\includegraphics[width=0.14\linewidth]{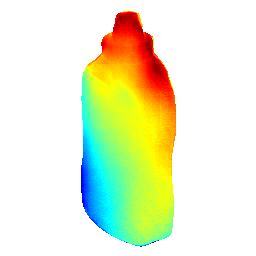}}&
  \raisebox{-0.5\height}[0pt][0pt]{
            \begin{tikzpicture}
                \node[opacity=1.0, anchor=south west, inner sep=0pt] at (0pt, 6pt)
                {\includegraphics[width=0.2cm, height=4cm]{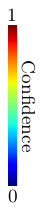}};
            \end{tikzpicture}
        }\\

  \setlength{\fboxsep}{-0.5pt}\setlength{\fboxrule}{0.5pt}\fbox{\includegraphics[width=0.14\linewidth]{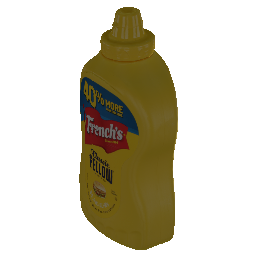}} &
  \setlength{\fboxsep}{-0.5pt}\setlength{\fboxrule}{0.5pt}\fbox{\includegraphics[width=0.14\linewidth]{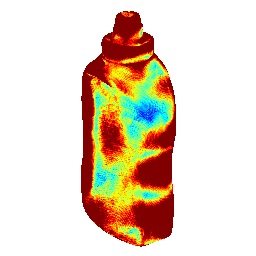}} &
  \setlength{\fboxsep}{-0.5pt}\setlength{\fboxrule}{0.5pt}\fbox{\includegraphics[width=0.14\linewidth]{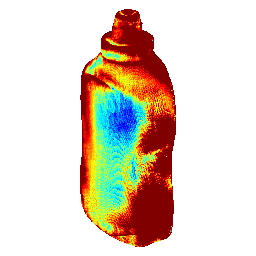}} &
  \setlength{\fboxsep}{-0.5pt}\setlength{\fboxrule}{0.5pt}\fbox{\includegraphics[width=0.14\linewidth]{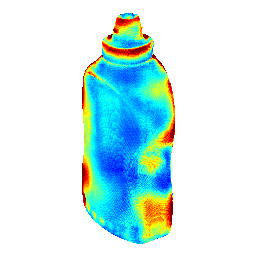}} &
  \setlength{\fboxsep}{-0.5pt}\setlength{\fboxrule}{0.5pt}\fbox{\includegraphics[width=0.14\linewidth]{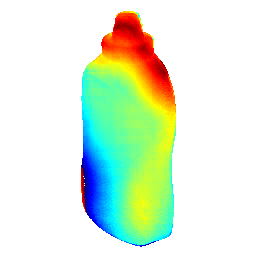}} &
  \setlength{\fboxsep}{-0.5pt}\setlength{\fboxrule}{0.5pt}\fbox{\includegraphics[width=0.14\linewidth]{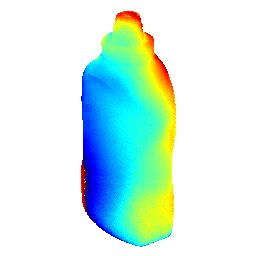}} &
  \setlength{\fboxsep}{-0.5pt}\setlength{\fboxrule}{0.5pt}\fbox{\includegraphics[width=0.14\linewidth]{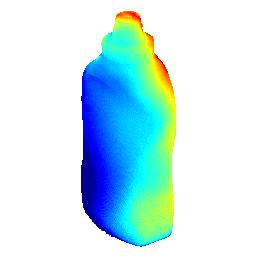}} &
  ~\\

   {\small Input} & \multicolumn{3}{c}{{\cellcolor{blue!10} {\small Variance $\sigma$}}}
   & \multicolumn{3}{c}{{\cellcolor{green!10} {\small Flow error}}}  & ~\\
\end{tabular}
}
    \caption{{\bf Visualization with increasing training iterations.}
    We show the normalized visualization of variance $\sigma$ and flow error at training steps 40k, 70k, and 100k from left to right, respectively. 
    The top row shows the baseline method~\cite{tarvainen2017mean}, and the bottom row is ours.
    The baseline method struggles to produce valid flow labels and reliable optical flow, regardless of the training steps. By contrast, our method produces more and more valid flow labels with small variance $\sigma$ during training, also with progressively better flow predictions.
    }
    \label{fig:training_sigma}
\end{figure*}

\noindent \textbf{Training analysis on YCB-V.}
We evaluate our method in three different settings during the training, as shown in Fig.~\ref{fig:training}. The baseline is the original teacher-student structure~\cite{tarvainen2017mean}, and the other two are the proposed components of our method. The baseline model struggles to learn from the unannotated data without explicit quality measurement of pseudo labels. Our flow loss introduces a constraint based on flow consistency derived from multiview geometry, and tackles this problem effectively, and the proposed photometric loss increases performance further, as shown in Fig.~\ref{fig:training_sigma}.

\begin{table}[]
    \centering

\begin{tabular}{cc|cccc}
   \toprule
   $\mathcal{L}_{flow}$            & $\mathcal{L}_{photo}$                 &  MSPD        & MSSD          & VSD                   & ADD\\
   \midrule
    -              & -                     &  0.759        &  0.589        & 0.519         & 30.9 \\
     -             & \cmark                &  0.765        &  0.631        & 0.578          & 37.5          \\
   \cmark          & -                     &  0.780        &  0.711        & 0.658          & 64.2   \\
   \cmark          & \cmark                &  {\bf 0.785}  & {\bf 0.749}   & {\bf 0.664}     & {\bf 67.4}\\
   \bottomrule
\end{tabular}
    \vspace{0.2em}
    \caption{{\bf Evaluation of different components on YCB-V.}
    $\mathcal{L}_{flow}$ is the key component of our framework, and $\mathcal{L}_{photo}$ improves the performance further.}
    \label{tab:compare_ssl}
\end{table}

\begin{table}
    \centering
    \scalebox{0.95}{

\begin{tabular}{l|ccccc}
   \toprule
   Method              & MSPD          & MSSD           & VSD          & ADD\\
   \midrule
   Initialization v1            & 0.632         & 0.491         & 0.420         & 27.4 \\
   + Ours (Real)  & {\bf 0.780}   & {\bf 0.731}   & {\bf 0.673}   & {\bf 64.6}       \\
   {\bf + Ours (SSL)}  & \underline{0.759}         & \underline{0.722}         & \underline{0.650}         & \underline{63.2}\\
   \midrule
    Initialization v2     & 0.673         & 0.580         & 0.508         & 36.0\\
   + Ours (Real)  & {\bf 0.775}   & \underline{0.722}         & {\bf 0.660}   & {\bf 65.3}    \\
   {\bf + Ours (SSL)}  & \underline{0.764}         & {\bf 0.724}   & \underline{0.643}         & \underline{64.2}\\
   \midrule
    Initialization v3          & 0.694         & 0.598         & 0.522         & 38.6  \\
   + Ours (Real)  & {\bf 0.803}   & {\bf 0.752}   & {\bf 0.686}   & {\bf 69.2}      \\
   {\bf + Ours (SSL)}  & \underline{0.785}         & \underline{0.749}         & \underline{0.664}         & \underline{67.4}\\
   
   \bottomrule
\end{tabular}
    }
    \vspace{0.2em}
    \caption{{\bf Performance with different initialization and additional annotations on YCB-V.}
We evaluate three versions of pose initialization with different accuracy, including the results obtained by the original WDR-Pose (``Initialization v1''), and also two other versions with pre-cropping the detected regions of interest, based on Mask RCNN and RADet, respectively (``v2'' and ``v3''). Our self-supervised refinement (``SSL'') boosts the initialization accuracy significantly and achieves similar performance as versions trained with fully-annotated real images (``Real'').
    }
    \label{tab:different_pose_init_ycbv}
\end{table}

\begin{figure*}[h]
    \centering
    \setlength\tabcolsep{1.5pt}
    \scalebox{0.97}{
\begin{tabular}{cccccc}
  \includegraphics[width=0.16\linewidth]{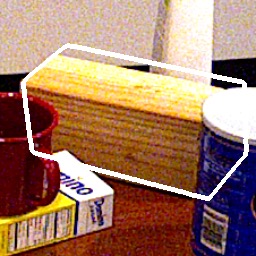} &
  \includegraphics[width=0.16\linewidth]{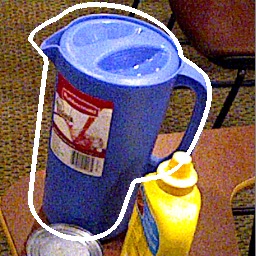} &
  \includegraphics[width=0.16\linewidth]{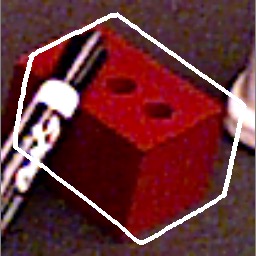} &
  \includegraphics[width=0.16\linewidth]{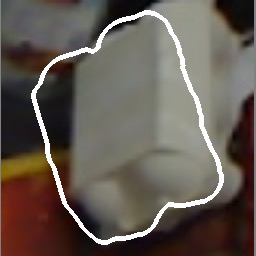} &
  \includegraphics[width=0.16\linewidth]{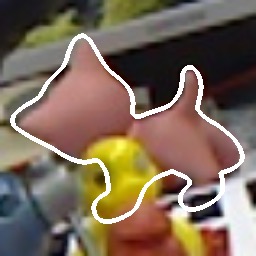} & 
  \includegraphics[width=0.16\linewidth]{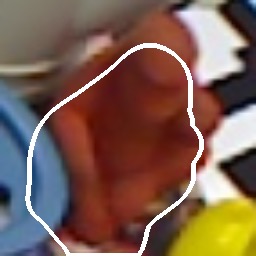} \\
  \includegraphics[width=0.16\linewidth]{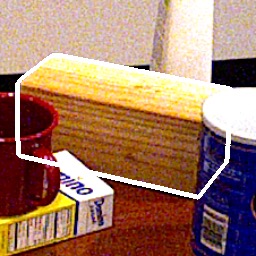}& 
  \includegraphics[width=0.16\linewidth]{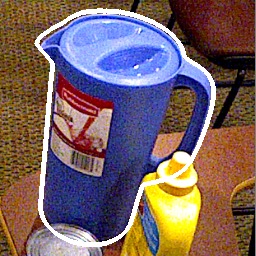} &
  \includegraphics[width=0.16\linewidth]{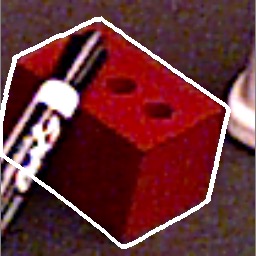} &
  \includegraphics[width=0.16\linewidth]{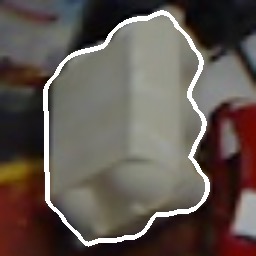}& 
  \includegraphics[width=0.16\linewidth]{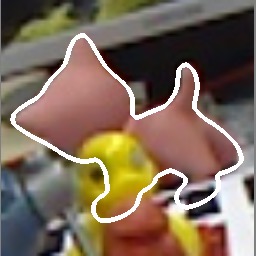}& 
  \includegraphics[width=0.16\linewidth]{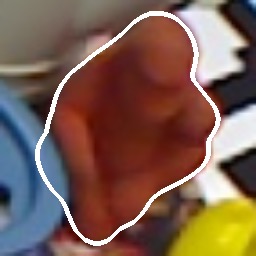}\\
\end{tabular}
}
    \caption{{\bf Qualitative results.}
    We show the initialization results trained only on synthetic data at the top, and the refinement results after using our self-supervised strategy at the bottom.
    Our method significantly improves the baseline in various conditions, such as occluded, weak-textured, and symmetry objects.
    }
    \label{fig:quantitive_reuslts}
\end{figure*}

\begin{figure*}
    \centering
    \setlength\tabcolsep{4pt}
    \begin{tabular}{ccc} 
        \includegraphics[width=0.3\linewidth]{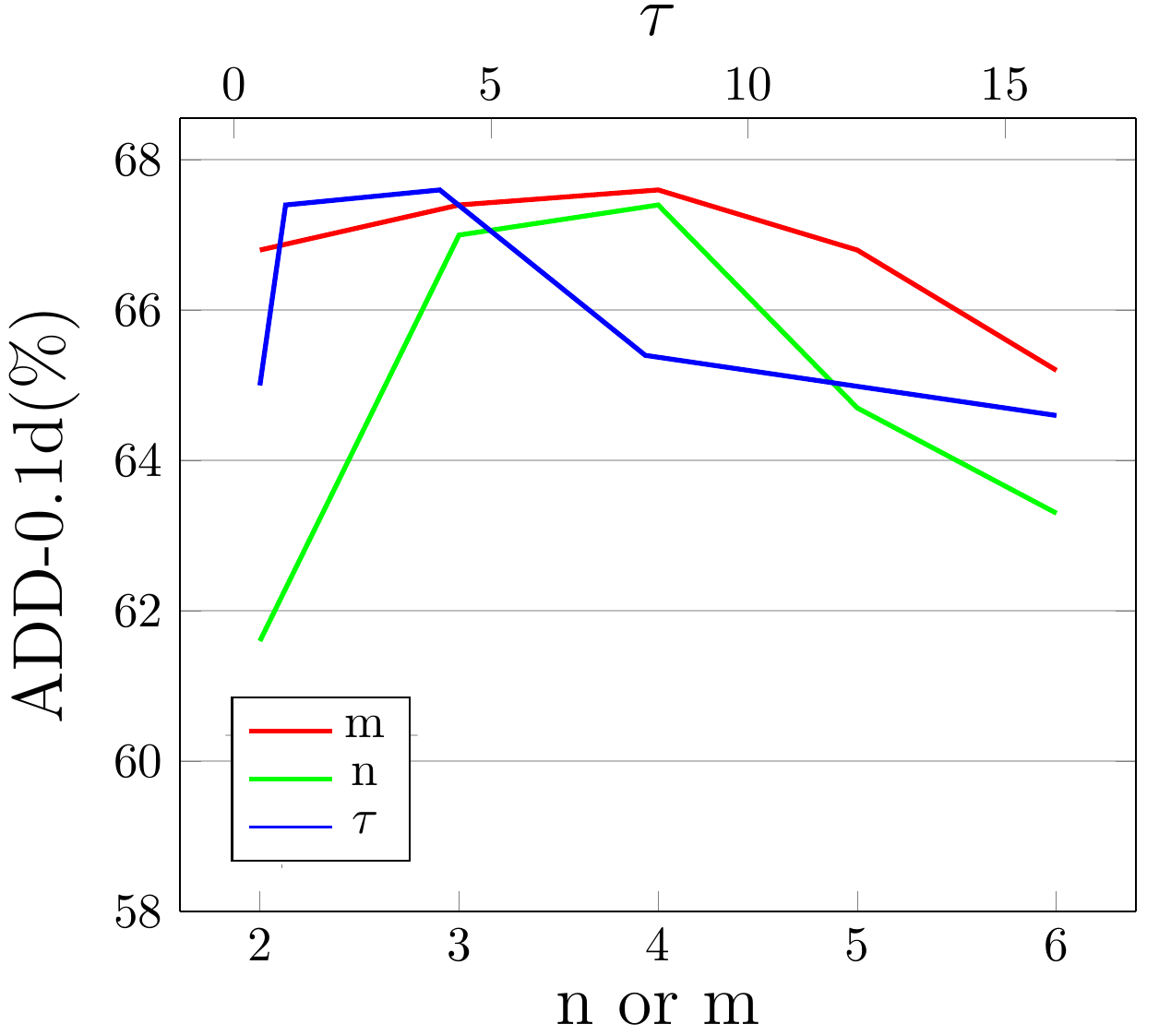} &
        \includegraphics[width=0.31\linewidth]{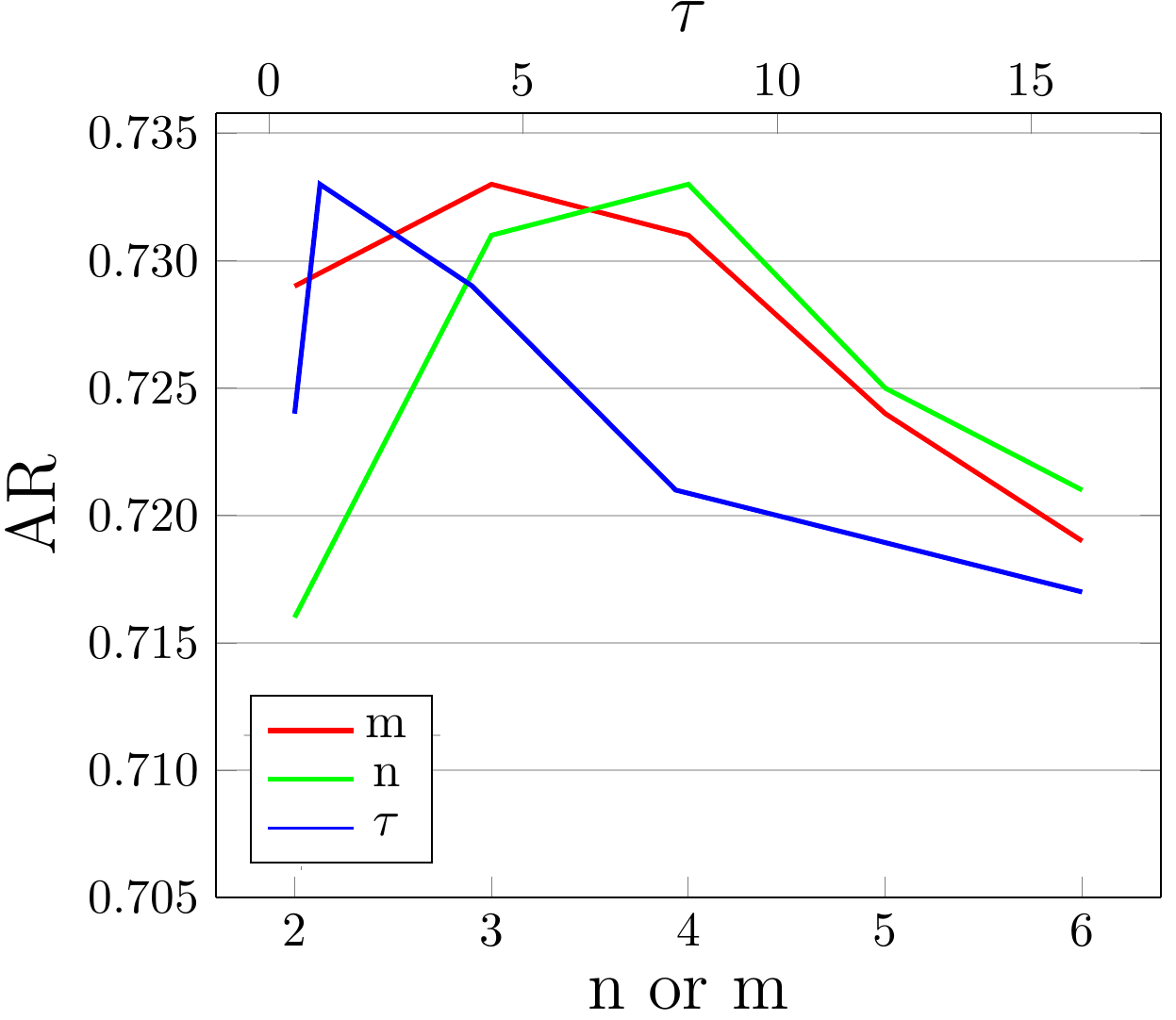} & 
        \includegraphics[width=0.29\linewidth]{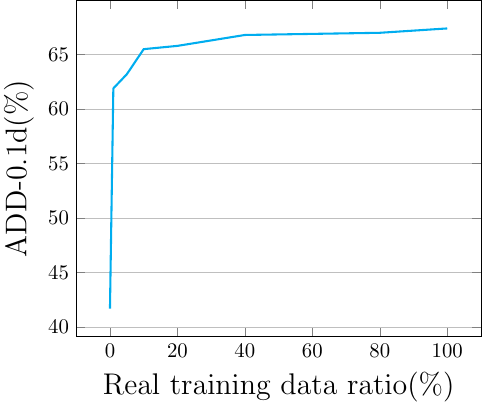} \\
        \multicolumn{2}{c}{\small (a) Ablation study in ADD-0.1d and AR} & {\small (b) Different amount of real training data}\\
    \end{tabular}
    \vspace{0.2em}
    \caption{{\bf Ablation study of hyper-parameters and training data on YCB-V.}
    {\bf (a)} $m$ and $n$ are the number of views used for $\mathcal{L}_{photo}$ and $\mathcal{L}_{flow}$, respectively, and $\tau$ is the threshold for determining the validity of flow labels for $\mathcal{L}_{flow}$. Our method is robust to the choices of different hyper parameters. {\bf (b)} Our method produces acceptable results with accessing only 1\% of all the real data.
    }
    \label{fig:ablate_total}
\end{figure*}

\noindent \textbf{Evaluation of hyper-parameters.}
We evaluate the hyper-parameters used in our framework in Fig.~\ref{fig:ablate_total}.
We first evaluate the impact of the number of different views.
More views generally increase the performance, since it adds more information to the geometry constraint. However, too many views, such as those larger than 4, has negative impacts on the performance. We believe it is caused by the noise introduced by too many views with large viewpoint differences, which usually makes the network harder to learn.
We then evaluate the threshold $\tau$ used in our framework, which is used to determine the reliability of pseudo flow labels. It works well between 1 and 4.

\noindent \textbf{Evaluation with limited real data.}
We evaluate our method on YCB-V with different amounts of real data used in training, as shown in Fig.~\ref{fig:ablate_total}(b). Our method increases the performance by about 20.2\% (from 41.7\% to 61.9\%) in ADD-0.1d by only using 1\% of all the real data. 99\% more real data can only increase the performance by 5.5\% further, which demonstrates the effectiveness of our method in data-limited scenarios.

\noindent \textbf{Evaluation with different initialization and additional annotations.}
In principle, our method can be trained with ground truth pose annotations easily, which is basically training the student network in a standard fully-supervised way. We evaluate our self-supervised method with the version trained with real pose annotations. At the same time, to evaluate the robustness of our framework to different pose initialization, we evaluate three versions of pose initialization with different initialization accuracy, including the results obtained by the original WDR-Pose, and two other versions with pre-cropping the detected regions of interest, based on Mask RCNN~\cite{he2017mask} and RADet~\cite{yang2023radet}, respectively.
Table~\ref{tab:different_pose_init_ycbv} summarizes the results.
Although the initial poses have different accuracy, our self-supervised refinement framework boosts their performance significantly, and even achieves similar performance as that trained on fully-annotated real images.

\noindent \textbf{Comparison against standard optical flow methods.}
In principle, one can use a self-supervised optical flow method to directly establish dense 2D-to-2D correspondence between the rendered image and the real image input, without any real pose annotations. However, we find that this strategy hardly can work, mainly due to the large domain gap between the rendered and real images, in which case the standard component of photometric comparison in self-supervised flow methods suffers. We evaluate a typical self-supervised optical flow method SMURF~\cite{stone2021smurf}, and report the results on YCB-V in Table~\ref{tab:compare_unsup}. Our self-supervised strategy suffers little from this domain gap problem and produces much more accurate pose results than SMURF.

\begin{table}[]
    \centering
    \begin{tabular}{ccccc}
    \toprule
    Method      & MSPD      & MSSD      & VSD       & ADD \\
    \midrule
    SMURF~\cite{stone2021smurf}       &  0.751   & 0.565     & 0.488     & 28.7\\
    {\bf Ours}   & {\bf 0.785}     & {\bf 0.749}     & {\bf 0.664}     & {\bf 67.4}  \\
    \bottomrule
\end{tabular}
    \vspace{0.2em}
    \caption{{\bf Comparison against standard optical flow methods on YCB-V.}
    The typical self-supervised optical flow method SMURF~\cite{stone2021smurf} suffers in producing 6D object poses without pose annotations.
    }
    \label{tab:compare_unsup}
\end{table}

\begin{figure}
    \centering
    \setlength\tabcolsep{1pt}
    \renewcommand{\arraystretch}{0.8}
    \begin{tabular}{cccc}
        \setlength{\fboxsep}{-0.5pt}\setlength{\fboxrule}{0.5pt}\fbox{\includegraphics[width=0.23\linewidth]{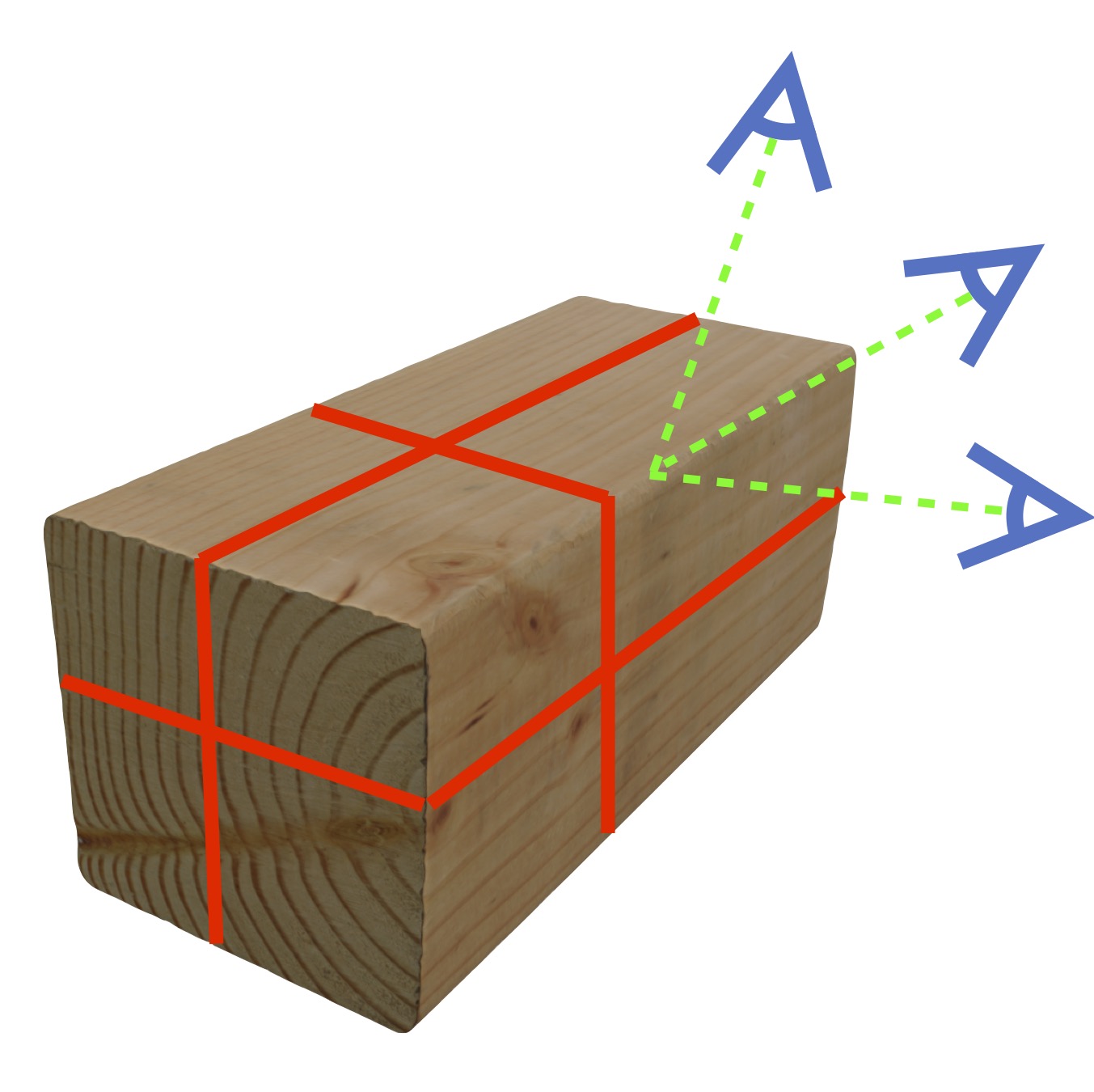}}&
        \includegraphics[width=0.23\linewidth]{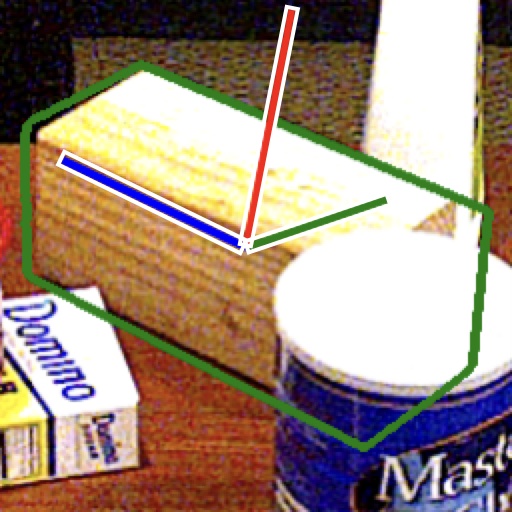} &
        \includegraphics[width=0.23\linewidth]{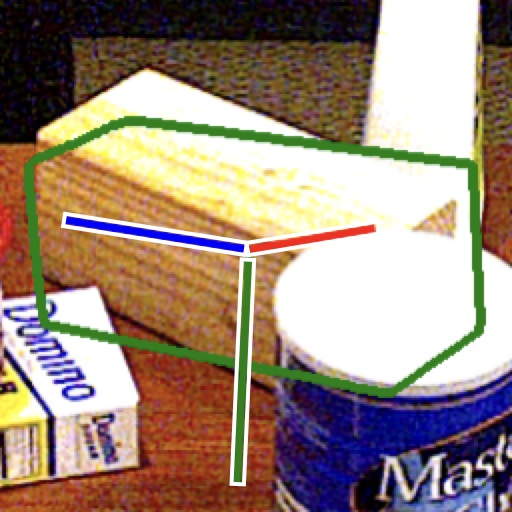} &
        \includegraphics[width=0.23\linewidth]{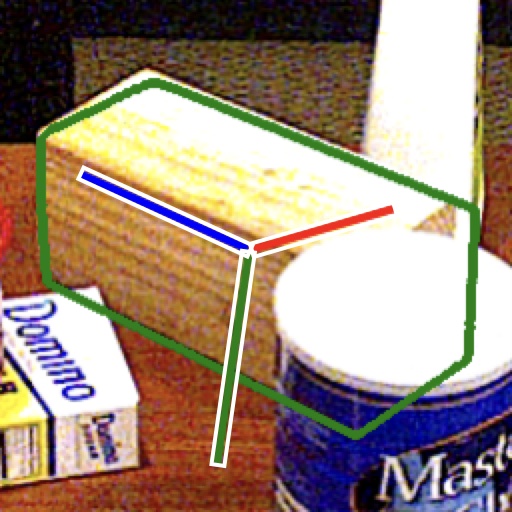} \\
        \setlength{\fboxsep}{-0.5pt}\setlength{\fboxrule}{0.5pt}\fbox{\includegraphics[width=0.23\linewidth]{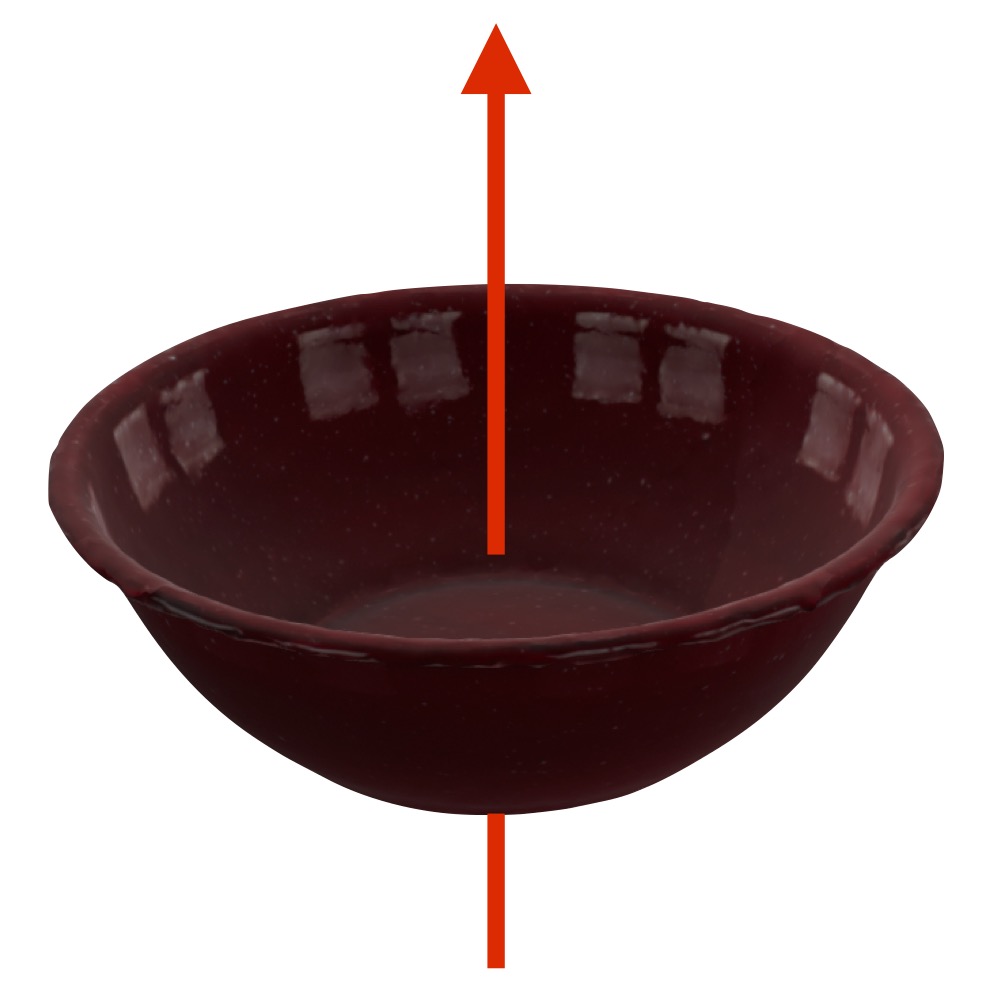}}&
        \includegraphics[width=0.23\linewidth]{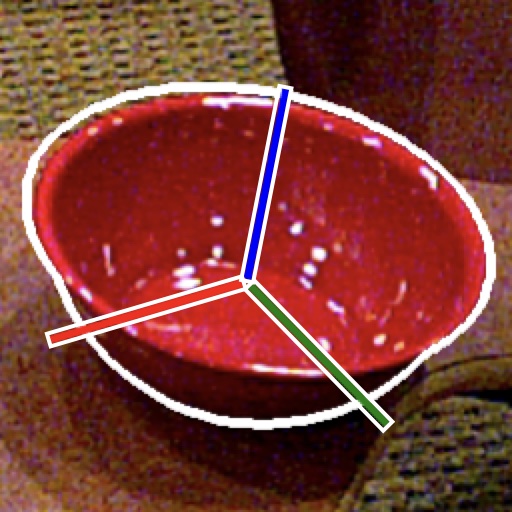} &
        \includegraphics[width=0.23\linewidth]{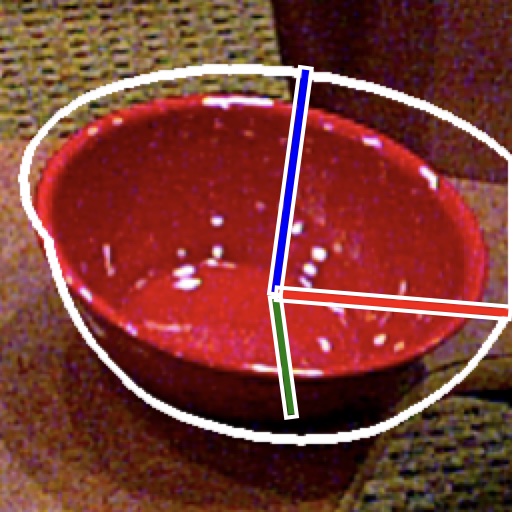} &  
        \includegraphics[width=0.23\linewidth]{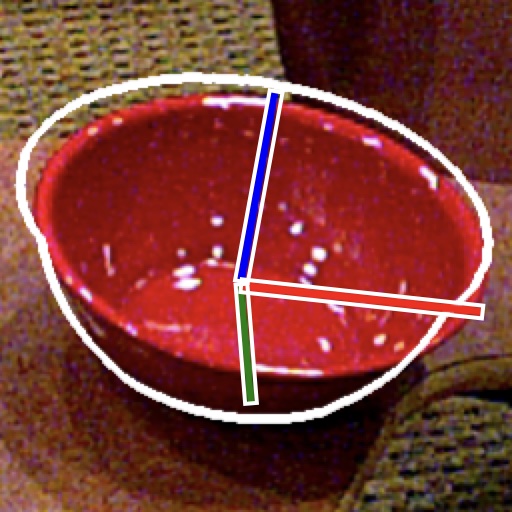} \\
        {\footnotesize Symmetry objects}   & {\footnotesize Ground truth} & {\footnotesize Initialization}  & {\footnotesize Refinement}  \\
    \end{tabular}
    \vspace{0.5em}
    \caption{{\bf Symmetric objects handling.} Our refinement results align the target mesh with the images well in appearance.
    }
    \label{fig:sym_handling}
\end{figure}

\noindent \textbf{Symmetry handling.}
We use the same strategy as in ~\cite{PFA_2022_eccv, rad2017bb8} for symmetric objects in obtaining pose initialization, which restricts the range of poses used in training depending on the objects' symmetry type. We do not explicitly handle the symmetry for the refinement, where we build multiple views around the initial pose and use the consistency constraint between them to find the best flow to align different views within a small pose range. Fig.~\ref{fig:sym_handling} shows two examples with reflectional symmetry and rotational symmetry, respectively. Note how the predicted poses have different 3D axis from the ground truth pose but align the target mesh with the images well in appearance.

\noindent \textbf{Time analysis.}
We conduct all our experiments on a workstation with an NVIDIA RTX-3090 GPU and an Intel-Xeon CPU with 12 cores running at 2.1GHz. 
The training of our self-supervised framework is only about 20\% slower than its fully-supervised version, consuming about 30 and 24 hours in the typical setting on YCB-V, respectively, which is much more efficient than methods relying on multiple times of retraining~\cite{lin2022learning, chen2022sim}.
For the inference time, our method is the same as its fully-supervised version and takes only 23ms for a single object, including the optical flow estimation 17ms and the PnP solver 6ms.

\section{Conclusion}

We have introduced a simple self-supervised 6D object pose method. After obtaining the rough pose initialization based on a network training on synthetic images, we refine the pose with a teacher-student pseudo labeling framework. To solve the problem of identifying high-quality labels in the context of object pose estimation, we first formulate pseudo object pose labels as pixel-level optical flow supervision signals. Then, we introduce a flow consistency based on the underlying geometry constraint between multiple different views. Our experiments show that the proposed method significantly outperforms existing solutions in both accuracy and efficiency, without relying on any 2D annotations or additional depth images.

\vspace{0.2em}
{\small
{\noindent \bf Acknowledgments.}
This work was supported by the Natural Science Foundation of China, the Youth Innovation Team of Shaanxi Universities, the Fundamental Research Funds for the Central Universities under Grant JBF220101, and the 111 Project of China under Grant B08038. We thank Thomas Probst, Daniel Koguciuk, Juan-Ting Lin, and Lin Sun for helpful discussions.
}

{\small
\bibliographystyle{ieee_fullname}
\bibliography{egbib}
}

\end{document}